\documentclass{article}

\usepackage{microtype}
\usepackage{graphicx}
\usepackage{subfigure}
\usepackage{booktabs} 

\usepackage{verbatim}
\usepackage{graphicx}
\usepackage[margin=1.2in]{geometry}
\usepackage{amsmath}
\usepackage{amssymb}
\usepackage{amsthm}
\usepackage{algorithm}
\usepackage{algorithmic}
\usepackage{wrapfig}
\usepackage{natbib}

\usepackage{hyperref}

\date{\vspace{-5ex}}

\title{Predict and Constrain: Modeling Cardinality in Deep Structured Prediction}

\author{
        Nataly Brukhim \\ \texttt{natalybr@mail.tau.ac.il}
            \and
       Amir Globerson \\
       \texttt{gamir@cs.tau.ac.il}
}

\begin{document}
\maketitle

\begin{abstract}
Many machine learning problems require the prediction of multi-dimensional labels. Such structured prediction models can benefit from modeling dependencies between labels. Recently, several deep learning approaches to structured prediction have been proposed. Here we focus on capturing cardinality constraints in such models. Namely, constraining the number of non-zero labels that the model outputs. Such constraints have proven very useful in previous structured prediction approaches, but it is a challenge to introduce them into a deep learning framework. Here we show how to do this via a novel deep architecture. Our approach outperforms strong baselines, achieving state-of-the-art results on multi-label classification benchmarks.
\end{abstract}

\section{Introduction}
\label{Introduction}
Deep structured prediction models have attracted considerable interest in recent years (\citealt{belanger2017end}; \citealt{zheng2015conditional}; \citealt{schwing2015fully}; \citealt{chen2015learning}; \citealt{ma2016end}). The goal in the structured prediction setting is to predict multiple labels simultaneously
while utilizing dependencies between labels to improve accuracy. Examples of such application domains include dependency parsing and part-of-speech tagging for natural language processing, as well as visual recognition tasks such as semantic image segmentation.

In the conventional structured prediction setting
a non-linear classifier is trained first, and its output is used to produce potentials for the structured
prediction model. Arguably, this piece-wise training is suboptimal as the classifier is learned while ignoring the dependencies between the predicted variables. However, when trained jointly their predictive power can be increased by utilizing complementary information.

Several recent approaches have proposed to combine the representational power of deep neural networks with the ability of structured prediction models to capture variable dependence, trained jointly in an end-to-end manner. This concept has been proven efficient in various applications (\citealt{zheng2015conditional}; \citealt{schwing2015fully}; \citealt{ma2016end}; \citealt{belanger2017end}), despite mostly utilizing only basic structured prediction models restricted to limited variable interactions, e.g., pairwise potentials. In this work, we seek to leverage a deep structured prediction model to incorporate higher order structural relations among labels. 
 
One of the most effective forms of label structure is cardinality (\citealt{tarlow2012fast}; \citealt{tarlow2010hop}; \citealt{milch2008lifted}; \citealt{swersky2012cardinality}; \citealt{gupta2007efficient}). Namely, the fact that the overall number of labels taking a specific value has a distribution specific to the domain. Such potentials naturally arise in natural language processing, where they can express a constraint on the number of occurrences of a part-of-speech, e.g., that each sentence contains at least one verb \cite{ganchev2010posterior}. In computer vision, a cardinality potential can encode a prior distribution over object sizes in an image. 

Although cardinality potentials have been very effective in many structured prediction works, they have not yet been successfully integrated into deep structured prediction frameworks. This is precisely the goal of our work.

The challenge in modeling cardinality in deep learning is that cardinality is essentially a combinatorial notion, and it is thus not clear how to integrate it into a differentiable model. Our proposal to achieve this goal is based on two key observations.

First, we note that learning to predict \textit{how many} labels are active for a given input is easier than predicting \textit{which} labels are active. Hence, we break our inference process into two complementary components: we first estimate the label set cardinality for a given input using a learned neural network, and then predict a label that satisfies this constraint.

The second observation is that constraining a label to satisfy a cardinality constraint can be approximated via projected gradient descent, where the cardinality constraint corresponds to a set of linear constraints. Thus, the overall label prediction architecture performs projected gradient descent in label space. Moreover, the above projection can be implemented via sorting, and results in an end-to-end differentiable architecture which can be directly optimized to maximize any performance measure (e.g., $F_1$, recall at $K$,  etc.). Importantly, with this approach cardinality can be naturally integrated with other higher-order scores, such as the global scores considered in \citealt{Belanger:2016:SPE:3045390.3045495}, and used in our model as well. 

Our work augments the simple form of unrolled optimization \cite{Belanger:2016:SPE:3045390.3045495} by extending the scope of its applicability beyond gradient descent. We formulate cardinality potentials as a constraint on the label variables, and demonstrate their capability of capturing important global structures of label dependencies. 

Our proposed method significantly improves prediction accuracy on several datasets, when compared to recent deep structured prediction methods. We also experiment with other
approaches for modeling cardinality in deep structured prediction, and observe that our Predict and Constrain method outperforms these. 

Taken together, our results demonstrate that deep structured prediction models can benefit from representing cardinality, and that a Predict and Constrain approach is an effective method for introducing cardinality in a differentiable end-to-end manner.

\section{Preliminaries}
We consider the setting of predicting a set of $L$ labels $\mathbf{y} = (y_i)_{i \in [L]}$, with $[L] = \{1, ...,L\}$, given an input $\mathbf{x} \in \mathcal{X}$. We assume $y_i$ are binary labels, however our approach can be generalized to the multi-class case. 
 
To model our problem in a structured prediction framework, we define a score function $s(\mathbf{x}, \mathbf{y})$ over the input-output pairs. In this setting, $s(\mathbf{x}, \mathbf{y})$ is learned to evaluate different input-output configurations such that its maximizing assignment over the label space approximates the ground-truth label $\mathbf{y}^*$. Thus, for a given input $\mathbf{x}$, we wish to obtain,
\begin{equation} \label{argmax_score}
\hat{\mathbf{y}} = \underset{\mathbf{y}}{\text{argmax }}s(\mathbf{x}, \mathbf{y})
\end{equation}
In practice, we will maximize over $y$ by relaxing the integrality constraint and using projected gradient descent. 
We assume that $s(\mathbf{x}, \mathbf{y})$ can be decomposed as follows,
\begin{equation} 
s(\mathbf{x}, \mathbf{y}) = \sum\limits_i s_i(\mathbf{x}, y_i) +  s_g(\mathbf{y}) +  s_{z(x)}(\mathbf{y})
\end{equation}

where $s_i$ is a function that depends on a single output variable (i.e., unary potential), $s_g$ is an arbitrary learned global potential defined over all variables and independent of the input, and $s_{z(x)}$ is a global cardinality potential which enforces the maximizing labeling to be of cardinality $z(x)$, as detailed in Section \ref{section3}. For brevity, we denote the cardinality potential as $s_z$.

Concretely, $s_i(\mathbf{x}, y_i)$ is defined by multiplying $y_i$ by a linear function of a feature representation of the input, where the feature representation is given by a multi-layer perceptron $f(\mathbf{x})$. The global score $s_g$ is obtained by applying a neural network over the variables, and evaluates such assignments independent of $\mathbf{x}$, similarly to the architecture used by \citealt{Belanger:2016:SPE:3045390.3045495}.

The score function $s(\mathbf{x}, \mathbf{y})$ is parameterized by a set of weights $\mathbf{w}$. During training we seek the value of these weights which minimizes a loss function over the predicted output $\mathbf{y}$ with respect to the ground-truth label $\mathbf{y}^*$. Recall that prediction is made by applying projected gradient descent over the relaxed variables. In what follows we describe our method in more detail.

\section{Learning with Cardinality Potentials} \label{section3}

We propose a deep structured architecture, with the objective of maximizing the score function $s(\mathbf{x}, \mathbf{y})$. The score function takes into account the independent relevance of each variable, as given by the unary potentials, as well as variable correlations modeled by the global potential. In addition, we enhance the ability of our model to represent complex structural relations between variables, utilizing the expressiveness of cardinality potentials. Such potentials are capable of capturing important label dependencies which are harder to model using an arbitrary form global potential. 

Specifically, we formulate the cardinality score as a constraint on the sum of labels defined as follows,
$$
s_z = 
\begin{cases}
0 & \text{if  } \sum_iy_i = z \\
- \infty & \text{otherwise  }
\end{cases}
$$
Such cardinality potentials are able to use the fact that distributions of label counts, pertaining to a specific value, are dependent on the task of interest. Moreover, we exploit the fact that $z$ can be predicted given $\mathbf{x}$, using a learned cardinality predictor $z = h(\mathbf{x})$. This enhances the power of cardinality potentials, by imposing a constraint on the output labels, tailored for their corresponding input data. 

Next, we wish to maximize the overall score function $s(\mathbf{x}, \mathbf{y})$ with respect to $\mathbf{y}$, taking into account both the unary and global potentials, as well as the cardinality potential. There are several ways to address this problem, detailed in Section \ref{review}, however we found our Predict and Constrain approach to be the most effective.

\begin{figure}[t]
\vskip 0.2in
\begin{center}
\centerline{\includegraphics[width=\columnwidth]{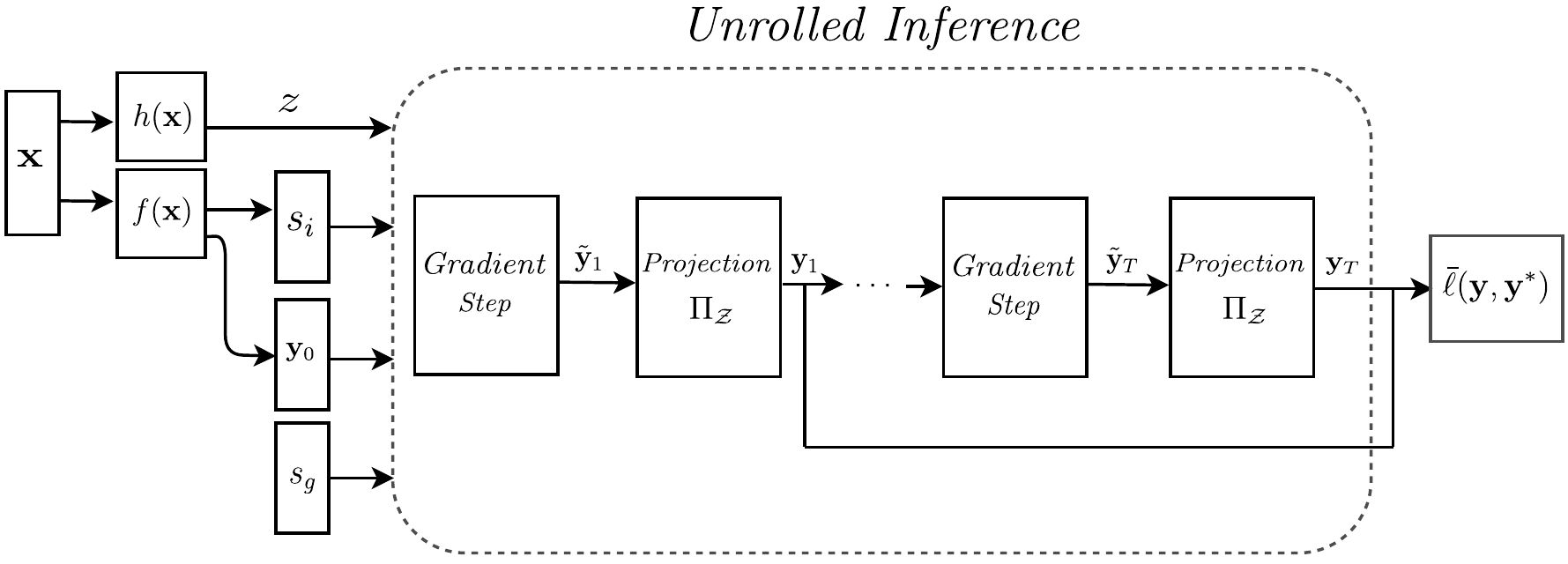}}
\vspace{-0.2mm}
\caption{Deep structured model with unrolled projected gradient descent inference.}
\vspace{-0.2mm}
\label{architecture_fig}
\end{center}
\vskip -0.2in
\end{figure}

\subsection{Learning Through Optimization}
Given input $\mathbf{x}$, we wish to obtain $\hat{\mathbf{y}}$, the maximizing labeling of the score function. However, as $s(\mathbf{x}, \mathbf{y})$ represents a complex non-linear function of $\mathbf{y}$, finding the maximizing label is intractable. Additionally, we are interested in constructing an end-to-end differentiable network, to take into account the inference process while learning the network parameters. 

We devise a differentiable approximation of $\hat{\mathbf{y}}$, by employing projected gradient descent as our inference scheme, as depicted in Figure \ref{architecture_fig}. To this end, we relax the discrete variables $\mathbf{y}$ to be in the interval $[0, 1]$. We unroll a gradient update step for $T$ iterations, which is essentially a sequence of differentiable updates to variables. Thus, gradients of the network parameters can backpropagate through the unrolled optimization in a similar manner to backpropagation through a recurrent neural network.

The detailed differentiation of unrolled gradient descent is given in \citealt{maclaurin2015gradient}, though in practice this computation is done using deep learning libraries for which the gradient components of the computation graph are differentiable, removing the need to explicitly program it. Similarly, the basic gradient update can be replaced by other differentiable optimization techniques. We leverage this idea to construct a differentiable architecture which computes a Euclidean projection following each gradient step.

Similar to other end-to-end approaches as \citealt{belanger2017end}, we found it useful to apply a loss function over all $T$ iterations. Specifically, we opted for the following weighing of the single-step loss terms, obtaining the overall loss function,

$$
\bar{\ell}(\mathbf{y}_0,...,\mathbf{y}_T, \mathbf{y}^*) = \frac{1}{T} \sum\limits_{t=1}^T  \frac{\ell(\mathbf{y}_t, \mathbf{y}^*)}{T -t +1}
$$

where $\ell(\mathbf{y}, \mathbf{y}^*)$ is a differentiable loss function defined for a single gradient step. By using this method the model learns to converge quickly during inference, and also diminishes the problem of vanishing gradients, as the loss function directly incorporates every layer. 

\subsection{Enforcing Cardinality via Projection} \label{projection_section}

During inference we iteratively apply projected gradient ascent over the variables for $T$ steps, where the initial variables $\mathbf{y}_0$ are given by a sigmoid function applied over the unary terms. 
In each step $t \in [T]$ of the inference pipeline we update the label variables as follows,
$$
\mathbf{y}_{t+1} = \Pi_{\mathcal{Z}}\big[\mathbf{y}_t + \eta \nabla_{\mathbf{y}} \bar{s}(\mathbf{x}, \mathbf{y})\big]
$$
where $\bar{s}(\mathbf{x}, \mathbf{y}) = \sum\limits_i s_i(y_i, \mathbf{x}) +  s_g(\mathbf{y})$, $\eta$ is the inference learning rate, and $\Pi_{\mathcal{Z}}$ denotes a projection  operator. The operator $\Pi_{\mathcal{Z}}$ differentially computes an approximation of a Euclidean projection onto a set defined as,
$$
\mathcal{Z} = \{\mathbf{y} | \forall i. y_i \in [0,1] \text{, } \sum_i y_i = z\}
$$
with $z$ obtained using the cardinality predictor $z = h(\mathbf{x})$. Thus, $\Pi_{\mathcal{Z}}(\mathbf{v})$ approximately solves the following problem,
\begin{equation} \label{proj_opt}
\begin{aligned}
& \underset{\mathbf{u}}{\text{minimize}}
& & \frac{1}{2}\left\lVert \mathbf{v} - \mathbf{u}\right\rVert_2^2
\\
& \text{subject to}
&& \sum_{i=1}^L u_i = z, \\ 
&&& 0 \leq u_i \leq 1, \; \forall i \in [L].
\end{aligned}
\end{equation}

First, we note that it is possible to compute the above minimization directly, however this is harder to obtain in our end-to-end differentiable network, as detailed in Section \ref{exact_proj}. Instead, we construct operator $\Pi_{\mathcal{Z}}$ using two sub-processes, each computing a projection onto a different set, $\mathcal{A}$ and $\mathcal{B}$, such that $\mathcal{Z} = \mathcal{A} \cap \mathcal{B}$. Let $\mathcal{A} = \{\mathbf{y} | \forall i. y_i \leq 1\}$. Given $\mathbf{y}$, its Euclidean projection onto  $\mathcal{A}$ is obtained simply by clipping values larger 
than $1$, i.e. $\Pi_{\mathcal{A}}(\mathbf{y}) = \text{min}(\mathbf{y}, 1)$. 
 Let $\mathcal{B} = \{ \mathbf{y} | \forall i. y_i \geq 0 \text{, } \sum_i y_i = z\}$, the positive simplex. When $z=1$, $\mathcal{B}$ is the probability simplex. The Euclidean projection onto $\mathcal{B}$ can be done using an $O(L\log L)$ algorithm which relies on sorting \cite{duchi2008efficient}. 
 
In designing our end-to-end differentiable network, we must take into account the smoothness restriction of our network components. Hence, we devise a differentiable variation of the simplex projection algorithm. The soft procedure for computing $\Pi_{\mathcal{B}}(\mathbf{y})$ is given in Algorithm \ref{alg1}. For differential sorting we use a sorting component based on a differential variation of Radix sort, built in the deep learning library TensorFlow \cite{tensorflow2015-whitepaper}. It is an interesting direction for future research to explore other differentiable sorting operations that can be computed more efficiently.

\begin{algorithm}[t]
\caption{Soft projection onto the simplex} 
\label{alg1}
\begin{algorithmic}
   \STATE {\bfseries Input:} vector $\mathbf{v} \in \mathbb{R}^L$, cardinality $z \in [Z]$
   \STATE Sort $\mathbf{v}$ into $\boldsymbol{\mu}$: $\mu_1 \geq ... \geq \mu_L$
   \STATE Compute $\bar{\boldsymbol{\mu}}$ the cumulative sum of $\boldsymbol{\mu}$
   \STATE Let $\mathbf{I}$ be a vector of indices $1,...,L$. 
   \STATE $\boldsymbol{\delta} = \text{Softsign}\big(\boldsymbol{\mu} \circ \mathbf{I} - (\bar{\boldsymbol{\mu}} - z) \big)$
   \STATE $\boldsymbol{\rho} = \text{Softmax}\big(\boldsymbol{\delta} \circ \mathbf{I}\big)$
   \STATE $\theta = \frac{1}{\mathbf{I}^{\text{T}} \boldsymbol{\rho}} \big(\bar{\boldsymbol{\mu}}^{\text{T}} \boldsymbol{\rho} - z\big)$
   \STATE {\bfseries Output:} $\text{max}(\mathbf{v} - \theta, 0)$
\end{algorithmic}
\end{algorithm}

Next, we need to combine the outputs of the operators $\Pi_{\mathcal{A}}(\mathbf{y})$ and $\Pi_{\mathcal{B}}(\mathbf{y})$ to output the desired $\Pi_{\mathcal{Z}}(\mathbf{y})$.
If $\mathcal{A}$ and $\mathcal{B}$ were affine sets we could have applied the alternating projection method \cite{Escalante:2011:APM:2077655}, by alternately projecting onto the sets $\mathcal{A}$ and $\mathcal{B}$. In this case, the method is guaranteed to converge to the Euclidean projection of $\mathbf{y}$ onto the intersection $\mathcal{Z} = \mathcal{A} \cap \mathcal{B}$. Since these are not affine sets due to the inequality constraints, this method is only guaranteed to converge to some point in the intersection. 

Instead, we use Dykstra's algorithm \cite{boyle1986method} which is a variant of the alternating projection method. Dykstra's algorithm converges to the Euclidean projection onto the intersection of convex sets, such as $\mathcal{A}$ and $\mathcal{B}$. 
Specifically, for each step $r \in [R]$ for a fixed number of iterations $R$  we compute the following sequence,
$$
\tilde{\mathbf{y}}^{(r)} = \Pi_{\mathcal{A}}(\mathbf{y}^{(r)} + \mathbf{p}^{(r)})
$$
$$
\mathbf{p}^{(r+1)} = \mathbf{y}^{(r)}  + \mathbf{p}^{(r)} - \tilde{\mathbf{y}}^{(r)}
$$
$$
\mathbf{y}^{(r+1)} = \Pi_{\mathcal{B}}(\tilde{\mathbf{y}}^{(r)} + \mathbf{q}^{(r)})
$$
$$
\mathbf{q}^{(r+1)} = \tilde{\mathbf{y}}^{(r)}  + \mathbf{q}^{(r)} - \mathbf{y}^{(r+1)} 
$$

Where $\mathbf{p}^{(0)} = \mathbf{q}^{(0)} = 0$. Empirically, we find that a small number of iterations is sufficient, and we set $R=2$ in all our experiments.

\section{Review of Alternative Approaches} \label{review}

In this section we review alternative inference methods that could be applied in our pipeline in place of our Predict and Constrain approach. Our goal is to further examine possible directions of dealing with complex global structures of the output labels, as well as to demonstrate the motivation behind the design choices made in our deep structured architecture. 

First, we consider an architecture which only consists of the unary and cardinality scores, $s_i$ and $s_z$, discarding the global potential $s_g$. In this case, exact maximization is possible by simply sorting the unary terms and obtaining the top $z$ values. The maximizer $\mathbf{y}^*$ is the binary vector in which the labels corresponding to the top $z$ values are on. This approach can be trained using standard structured hinge loss. Although appealing in its simplicity, this method fails to perform as well without utilizing the expressiveness of the global score $s_g$, which captures variable interactions that cannot be modeled by cardinality or unary potentials alone. SPENs \cite{Belanger:2016:SPE:3045390.3045495} have demonstrated the expressive power of such global scores to captures important structural dependencies among labels, such as mutual exclusivity and implicature.  

It is also possible to abandon constrained optimization altogether and instead of optimizing for variables $\mathbf{y} \in [0,1]$, we could optimize for the logits $\boldsymbol{\alpha}$, such that $\mathbf{y} = \sigma(\boldsymbol{\alpha})$, by applying the following gradient step,
$$
\boldsymbol{\alpha}_{t+1} = \boldsymbol{\alpha}_t + \eta \nabla_{\boldsymbol{\alpha}} s(\mathbf{x}, \sigma(\boldsymbol{\alpha}))
$$
where $\sigma(\cdot)$ is the sigmoid function. This tackles our ability to efficiently project onto the cardinality constrained space, which is a core part of our method. We could instead
design a global score $s_g$ in the hope it would be expressive enough to capture cardinality relations. However, such a neural network would require using a deeper structure with more parameters, and is thus prone to overfitting. A similar architecture was used by \citealt{belanger2017end}, which did not yield satisfactory results in our experiments, as demonstrated in Section \ref{experiments}.

Alternatively, we can replace the projection component of our pipeline with an architecture in which the cardinality potential is designed as a weighted sum of cardinality indicators. 
To this end, we can use the term $I_z(\mathbf{y}) = \sigma\big( \sum_i y_i - z \big)$ such that for label sets whose sum is at least $z$, $I_z$ is close to $1$, and to $0$ otherwise. Thus, we have, 
$$ 
s_z(\mathbf{y}) = \sum_{z \in [Z]} w_z \cdot I_z(\mathbf{y}) \cdot \big(1-I_{z+1}(\mathbf{y})\big)
$$
By using this method, the projection scheme collapses to the simple operation of clipping the values to be in the range $[0,1]$. Then, $s_z$ can be maximized along with the global and unary potentials using gradient-based inference, applying a clipping projection following each gradient step. 
However, this method does not encourage the inner optimization to enforce the cardinality constraint as much as directly projecting the variables onto the constrained space. In practice we found this method to underperform compared to our approach, as shown in Section \ref{experiments}.

Finally, it is possible to frame the task of maximizing $s(\mathbf{x}, \mathbf{y})$ as a mixed integer linear program, obtaining an exact inference scheme for our network. In order to train our model we could use the standard structured hinge loss. This formulation could also be relaxed to a linear program making it more efficient to solve. However, solving an LP for each prediction is impractical.

\subsection{Fast Exact Projection} \label{exact_proj}

Our projection scheme relies on Dykstra's algorithm which requires us to alternately apply $\Pi_{\mathcal{A}}$ and $\Pi_{\mathcal{B}}$ in an iterative fashion. In order to solve optimization problem \ref{proj_opt} to optimality we would need to encode several iterates of these alternations in our computation graph, to form a deeper network. Instead, we could compute the projection $\Pi_{\mathcal{Z}}$ by solving optimization problem \ref{proj_opt} directly, without separately applying $\Pi_{\mathcal{A}}$ and $\Pi_{\mathcal{B}}$. A solution to problem in Equation \ref{proj_opt} was given by \citealt{gupta2010l1}. They also describe a fast linear-time algorithm to obtain the maximizer $\mathbf{u}^*$. We will give a brief description of their method. 

Assume w.l.o.g. that $\mathbf{v}$ is sorted such that $v_1 \geq ... \geq v_n$. Let $\rho_1, \rho_2$ be the indices up until which all projected values are ones, 
and after which all projected values are zeros, respectively. Then, the values of the maximizer $\mathbf{u}^*$ take the following form,
\begin{equation} \label{u_maximizer}
u_i = 
\begin{cases}
0 & \text{if  } v_i - \lambda \leq 0 \\
v_i - \lambda & \text{if  } 0 < v_i - \lambda < 1 \\
1 & \text{if  }  v_i - \lambda \geq 1
\end{cases}
\end{equation}

with $\lambda$ defined as follows,
\begin{equation} \label{lambda}
\lambda = \frac{\sum_{i={\rho_1 + 1}}^{\rho_2} v_i - (z- \rho_1)}{\rho_2 - \rho_1}
\end{equation}

The algorithm suggested in \citealt{gupta2010l1} computes $\lambda$  in an iterative fashion, based on the fact that $z$ is a piece-wise linear function in $\lambda$ with points of discontinuity at 
values $\mathbf{v}$ and $\tilde{\mathbf{v}} = \text{max}(\mathbf{v} - 1, 0)$. The algorithm requires maintaining an uncertainty interval for $\lambda$, which is initialized at $[\text{min}(\tilde{\mathbf{v}}),\text{max}(\mathbf{v})]$. In each iteration $j \in [J]$ we obtain $\lambda_j$, the median of the merged set of unique values of $\tilde{\mathbf{v}}$ and $\mathbf{v}$, lying in the current uncertainty interval. We then need to compare $z_j$ to $z$, where $z_j$ can be computed using Equation \ref{lambda}. The size of the uncertainty interval is reduced in every iteration, until the correct value is recovered. 

Although this method efficiently computes the correct projection, it requires applying non-trivial combinatorial operations which do not naturally translate to differentiable operations, such as set-union, and median. We have experimented with a differentiable implementation of this algorithm to be used in our end-to-end network, detailed in Section \ref{experiments}. 

The projection algorithm requires many components added to the computation graph making it deeper and thus harder to backpropagate through. Alternatively, by iterating for only a small number of iterations, the resulting $\lambda_J$ is far from the correct $\lambda$. Overall, in this end-to-end setting, we found it to have inferior performance compared to the alternating projection method described in Section \ref{projection_section}. Instead, the use of Dykstra's algorithm with a few alternating projection iterations, was both efficient and thus easier to differentiate through, and obtained good a approximation of the correct maximizer. 

\section{Related Work}
Several recent approaches have applied gradient-based inference to a variety of structured prediction tasks (\citealt{Belanger:2016:SPE:3045390.3045495}; \citealt{gygli2017deep}; \citealt{amos2017input}). Specifically, Structured Prediction Energy Networks (SPENs) \cite{Belanger:2016:SPE:3045390.3045495} optimize the sum of local unary potentials and a global potential, trained with a structured SVM loss. Another approach is the Deep Value Network (DVN) \cite{gygli2017deep} which uses an energy network architecture similar to SPEN, but instead trains it to fit the task cost function. 

Our architecture was constructed similarly to SPEN and DVN for the unary and global potentials $s_i$, and $s_g$, though we extended the expressivity of this architecture both by introducing the cardinality potential $s_z$, as well as an effective inference method for the overall score.

Input Convex Neural Networks (ICNNs) \cite{amos2017input} design potentials which are convex with respect to the labels, so that inference optimization will be able to reach global optimum. Their design achieves convexity by restricting the model parameters and activation functions, limiting the expressiveness of the learned potentials. In practice, it has inferior performance compared to its non-convex counterpart, as shown by \citealt{belanger2017end}. 

Our approach differs from these methods by two main aspects. First, we embed the inference process within the learning model, and second, our model is capable of encapsulating complex global dependencies in the form of cardinality potentials, which are harder to obtain using general form global potentials.

\subsection{Cardinality Potentials}
The use of higher-order global potential, and specifically of cardinality relations, have shown to be useful in a wide range of applications. For example, in computer vision they have been used to improve human activity recognition \cite{hajimirsadeghi2015visual} by considering the number of people involved in an activity, which is harder to infer using spatial relations alone. In part-of-speech tagging, cardinalities can enforce the constraint that each sentence must contain at least one verb \cite{ganchev2010posterior}.

The properties of cardinality potentials and corresponding inference methods have been studied in a collection of prior work (\citealt{tarlow2012fast}; \citealt{tarlow2010hop}; \citealt{milch2008lifted}; \citealt{swersky2012cardinality}; \citealt{gupta2007efficient}). General form global potentials often result in non-trivial dependencies between variables that make exact inference intractable, thus requiring the use of approximate inference methods. Conversely, MAP inference for cardinality potential models is well-understood. Notably, \citealt{gupta2007efficient} shows an exact MAP inference algorithm, and \citealt{tarlow2010hop} gives a an algorithm for computing the cardinality potential messages for max-product belief propagation, both algorithms are solved efficiently in $O(L\text{log }L)$ time. Still, when considering general form global potentials as in our framework, we must employ an approximate inference scheme which can be computed efficiently. 

\subsection{Unrolled Optimization}
End-to-end training with unrolled optimization was first used in deep networks by \citealt{maclaurin2015gradient} for tuning hyperparameters. More recently, other approaches have unrolled gradient-based methods within deep networks in various different contexts (\citealt{metz2016unrolled}; \citealt{andrychowicz2016learning}; \citealt{greff2016highway}). 

In the context of inference, \citealt{belanger2017end} explored SPENs which use gradient descent to approximate energy minimization, while learning the energy function end-to-end. They have demonstrated that using unrolled optimization as an inference method can outperform baseline models that can be exactly optimized. In computer vision, several works have incorporated structured prediction methods like conditional random fields within neural networks (\citealt{zheng2015conditional}; \citealt{schwing2015fully}), where the Mean-Field algorithm is being used for inference. 

An important advantage of these training schemes is that they return not only the learned potentials, but also an actual inference optimization method, tuned on the training data, to be used at test time. However, these methods are either restricted to basic graphical models (e.g with pairwise or low order clique potentials) to ensure tractability, or have used global potentials of arbitrary form which are limited in their ability to capture interesting properties of the output space. Our approach harnesses the effectiveness of unrolled optimization while boosting its ability to infer important structures expressed by cardinality potentials.

\section{Experiments} \label{experiments}
We evaluate our method on multi-label classification datasets, for which the task is predicting a set of binary labels from text inputs given in a bag-of-words representation. The MLC task is relevant in a wide range of machine learning applications, and characterized by higher-order labels interaction, which can be addressed by our deep structured network. Therefore, it is a natural application of our method. We use 3 standard MLC benchmarks, as used by other recent approaches (\citealt{Belanger:2016:SPE:3045390.3045495}; \citealt{gygli2017deep}; \citealt{amos2017input}): Bibtex, Delicious, and Bookmarks. 

\subsection{Cardinality Prediction Analysis}
We begin by giving an analysis which demonstrates the effectiveness of estimating the cardinality of a label set given the input data. We train a simple feed-forward neural network which consists of a single hidden layer with ReLU activations, with the goal of predicting the ground-truth cardinality. The output layer is a softmax over $Z$ output neurons, where $Z$ is allowed the maximal cardinality. This is the same architecture we used for $h(\mathbf{x})$ in the larger setting, while in this experiment the data is the set of inputs $\mathbf{x}$ and their respective label cardinalities $|\mathbf{y}|$. 

We evaluate the results over the Delicious dataset using the mean squared error of our predictor's output with respect to the correct cardinality. We compare our predictor to a random baseline over the range of $[0,25]$, which is the range of possible cardinalities in the data, as well as to the constant cardinality of $19$ which is the average cardinality in the training data. Our predictor performs better than both baselines, with $\text{MSE}_{\text{rand}} = 74.9$, $\text{MSE}_{\text{const}} = 26.9$, and $\text{MSE}_h = 19.5$. 

A possible explanation for this phenomenon is that some attributes of the input data might indicate approximately how many active labels the label set contains. Features such as the number of distinct words, or the existence or absence of specific meaningful words could be relevant here, making the task of inferring \textit{how many} labels are active easier than \textit{which} labels are active. For example, an article with many distinct words suggests that it discusses a broad range of subjects, and thus relates to many different tags, while an article with few distinct words is more likely to be focused on a specific subject and therefore has only a small set of tags. Learning which combination of input words corresponds to which specific tagging set is harder than learning to predict cardinality based on feature representations of simpler forms, such as the ones discussed above.

The other datasets we tested, Bibtex and Bookmarks, are extremely sparse with average cardinalities of $2.4$ and $2$, respectively. The task of predicting the correct cardinality within this smaller possible range is harder. Accordingly, the predictor $h(\mathbf{x})$ approximately learns the average cardinality, and we observed in our experiments that projecting to a predicted cardinality $z = h(\mathbf{x})$ yields similar performance to projection onto a set defined by constant cardinality $z$. However, using the predictor $h(\mathbf{x})$ did manage to improve our overall performance for the Bibtex dataset compared to a fixed $z$, whereas for Bookmarks our results are slightly lower than for using a fixed $z$. 

The Predict and Constrain approach of first estimating the relevant cardinality and then projecting onto the cardinality-constrained space is especially useful for datasets of larger average cardinality and cardinality variance, such as Delicious, for which we obtained a significant performance increase using this method.

\subsection{Experimental Setup}
The evaluation metric for the MLC task is the macro-averaged $F_1$ measure. We found it useful to use its continuous extension as our loss function $\ell(\mathbf{y}, \mathbf{y}^*)$, i.e.,
\begin{equation} \label{f1_loss}
\ell(\mathbf{y}, \mathbf{y}^*) = -\frac{2\mathbf{y}^{\text{T}} \mathbf{y}^*}{\sum_i (y_i + y_i^*)}
\end{equation}
where $\mathbf{y}^*$ is binary and $\mathbf{y} \in [0,1]$. We observed improved performance by training with $\ell(\mathbf{y}, \mathbf{y}^*)$ as opposed to alternative loss functions, e.g. cross entropy.

The architecture used for all datasets consists of neural networks for the unary potentials $s_i$, the global potential $s_g$, and the cardinality estimator $h(\mathbf{x})$, respectively. For all neural networks we use a single hidden layer, with ReLU activations. For the unrolled optimization we used gradient ascent with momentum $0.9$, unrolled for $T$ iterations, with $T$ ranging between $10-20$, and with $R=2$ alternating projection iterations. All of the hyperparameters were tuned on development data. We trained our network using AdaGrad \cite{duchi2011adaptive} with learning rate $\eta = 0.1$.

We compare our method against the following baseline,
\begin{itemize}
  \item SPEN - Structured Prediction Energy Network \cite{Belanger:2016:SPE:3045390.3045495} uses gradient-based inference to optimize an energy network of local and global potentials, trained with SSVM loss. 
  \item E2E-SPEN - an end-to-end version of SPEN \cite{belanger2017end}.
  \item DVN - Deep Value Network \cite{gygli2017deep} trains an energy function to estimate the task loss on different labels for a given input, with gradient-based inference. 
  \item MLP - a multi-layer perceptron with ReLU activations trained with cross-entropy loss function.
\end{itemize}

The MLP and SPEN baseline results were taken from \cite{Belanger:2016:SPE:3045390.3045495}. The E2E-SPEN results were obtained by running their publicly available code on these datasets. In our experiments we follow the same train and test split as the baseline methods on all datasets. The results are shown in table \ref{results}.

\begin{table}[t]
\caption{$F_1$ performance of our approach compared to the state-of-the-art on multi-label classification.}
\label{results}
\vskip 0.15in
\begin{center}
\begin{small}
\begin{sc}
\begin{tabular}{lcccr}
\toprule
Dataset & Bibtex & Bookmarks & Delicious \\
\midrule
SPEN      & 42.2 & 34.4& 37.5\\
E2E-SPEN  & 38.1 & 33.9& 34.4\\
DVN       & 44.7 & 37.1&  - \\
MLP       & 38.9 & 33.8& 37.8\\
SC        & 42.0 & 34.6&  34.6\\ 
FP        & 42.1 & 36.0& 34.2\\
Ours      & $\boldsymbol{45.5}$ & $\boldsymbol{39.1}$& $\boldsymbol{38.4}$\\
\bottomrule
\end{tabular}
\end{sc}
\end{small}
\end{center}
\vskip -0.1in
\end{table}

\subsection{Alternative Implementations}
In additional to prior work, we have also compared our approach to methods that replace the cardinality enforcing component of our network, detailed in Section \ref{review}, to examine the role of our unrolled projection scheme in capturing cardinality relations. Specifically, we consider the following methods,
\begin{itemize}
  \item SC - The sigmoid-based indicators as cardinality potentials, discussed in Section \ref{review}. 
  \item FP - The fast exact projection method described in Section \ref{exact_proj}.
\end{itemize}

Both methods were trained with unrolled gradient ascent inference, using the negative $F_1$ loss function, shown in Equation \ref{f1_loss}.  

The fast projection method, was implemented using a differentiable approximation of the projection algorithm steps. The algorithm performs a binary search over the values of $z$ to obtain the correct $\lambda$, using set-union and median operations. Instead, we maintain a lower and upper bound for the uncertainty interval, and in each iteration we compute the average, rather than the median, until the gap between the lower and upper bounds is below a threshold. In each iteration $j$, we compared $z$ to $z_j$, given by Equation \ref{lambda}.

To obtain the values $\rho_1$, $\rho_2$, we compute $\boldsymbol{\rho}_1$ and $\boldsymbol{\rho}_2$, the one-hot encodings of $\rho_1$, $\rho_2$, in every iteration $j$. Let $\boldsymbol{\mu}$ be the sorted values of the vector we wish to project. We first compute $\boldsymbol{\delta}_1^{(j)}  = \text{Softsign}(\boldsymbol{\mu} - \lambda^{(j)} - 1)$ and $\boldsymbol{\delta}_2^{(j)}  = \text{Softsign}(\boldsymbol{\mu} - \lambda^{(j)})$. Then, we apply Softmax over the element-wise multiplication of a fixed indices vector and $\boldsymbol{\delta}_1^{(j)}$ or $\boldsymbol{\delta}_2^{(j)}$, to obtain $\boldsymbol{\rho}_1$ and $\boldsymbol{\rho}_2$, respectively. We compute the dot-product of the one-hot encodings and the cumulative sum vector of $\boldsymbol{\mu}$ to obtain the partial sum in Equation \ref{lambda}. Finally, we use Equation \ref{u_maximizer} to obtain the projected values.

\subsection{Discussion}

It can be seen from Table \ref{results} that our method outperforms all baselines we have compared against, obtaining state-of-the-art results in these tasks. Comparing our network to another end-to-end gradient-based inference method, the E2E-SPEN achieves significantly low performance. As stated by the authors in their release notes, their method is prone to overfitting and actually performs worse than the original SPEN on these benchmarks. Additionally, our method improves upon the original SPEN by a large margin. While SPENs obtain these results using pre-training of their unary potentials independently, our method is trained jointly without the need to pre-train any of our architecture components, in an end-to-end manner.

The DVN method underperformed on the Bibtex and Bookmarks datasets, compared to our method. They did not report results over the Delicious dataset, whereas running the code released by the authors yields extremely low results for Delicious. This suggests that further fine tuning of their method is required for different datasets. The Delicious dataset has the largest label space with $982$ labels, while Bibtex and Bookmarks have $159$ and $208$, respectively, and is therefore the most challenging. Thus, these results illustrate the robustness of our method, as it achieves superior performance for Delicious over all baselines, albeit not being specifically tuned for it. 

The performance of the SC and FP baselines is surpassed by that of our network. Nevertheless, they obtain competitive results compared to strong baselines. This further demonstrates the effectiveness of  unrolled optimization over applying an inference scheme separately from the training process. Moreover, these results indicate the power of combining general form global potentials with cardinality-based potentials to represent complex structural relations. 

By using an architecture similar to all examined baseline methods, for the unary and global scores, our results are demonstrative of the performance improvements obtained by enforcing cardinality constraints through the Predict and Constrain technique, over all three datasets.

\section{Extension to Non-Binary Labels} \label{non-binary}
Using the Predict and Constrain method in the multi-class case is an interesting extension of this work. For example, in image segmentation tasks in which each pixel could be assigned to one of multiple classes, cardinality potentials can enforce constraints on the size of labeled objects and encourage smoothness over large groups of pixels. 

To re-frame our method in the general non-binary form, we consider a matrix $\mathbf{Y} \in \{0,1\}^{L \times M}$ where $M$ is the number of possible classes, and $L$ the number of labels. Here, every row corresponds to the one-hot representation of the class $j \in [M]$ which label $i \in [L]$ is assigned to. We relax the matrix values to lie in the continuous interval $\mathbf{Y} \in [0,1]^{L \times M}$. We want to enforce the constraint that the rows of $\mathbf{Y}$ must lie in the probabilistic simplex, while the columns should obey cardinality constraints. We denote the space of matrices for which these constraints hold as $\tilde{\mathcal{Z}}$. 

Our approach translates into projection onto the intersection of two sets $\mathcal{C}$ and $\mathcal{D}$, such that $\tilde{\mathcal{Z}} = \mathcal{C} \cap \mathcal{D}$, where,
$$
\mathcal{C} = \{ \mathbf{Y} | \forall i,j. Y_{i,j} \geq 0 \text{, } \forall i. \sum_j Y_{i,j} = 1\}$$
$$\mathcal{D} = \{ \mathbf{Y} | \forall i,j. Y_{i,j} \geq 0 \text{, } \forall j. \sum_i Y_{i,j} = z_j\}$$ 

with $z_j$ being the cardinality constraint of class $j$. Since $\mathcal{C}$ and $\mathcal{D}$ are convex, we can again apply Dykstra's algorithm, alternating  between the projections onto the row constraints and column constraints. 

Since there is no dependence between the rows within $\mathcal{C}$, or columns within $\mathcal{D}$, the projections can be applied in parallel. Specifically, we can alternately apply simultaneous projection of all rows onto the unit simplex, and of all columns onto the positive simplex (using Algorithm \ref{alg1} with $z=1$ for each row, $z_j$ for each column $j$). We note that it is also possible to discard the inequality constraint in $\mathcal{D}$ to obtain a simpler projection algorithm.  

For cardinality prediction, we could learn a predictor $h_j(\mathbf{x})$ for each class $j$, so that we can then apply the projection operator of $\mathcal{D}$ for 
$z_j =  h_j(\mathbf{x})$. Thus, we utilize the benefits of first predicting the desired cardinalities of each class for a given input and then projecting the columns of $\mathbf{Y}$ onto the cardinality constrained space, which have have shown successful in our experiments for the binary label case.  

This extension of our framework generalizes our approach to be useful in a wide range of application domains. However, we did not experiment with this method, and it remains an interesting direction for future work.

\section{Conclusion}
This paper presents a method for using highly non-linear score functions for structured prediction augmented with cardinality constraints. We show how this can be done in an end to end manner, by using the algorithmic form of projection into the linear constraints that restrict cardinality. This results in a powerful new structured prediction model, that can capture elaborate dependencies between the labels, and can be trained to optimize model accuracy directly.

We evaluate our method on standard datasets in multi-label classification. Our experiments demonstrate that our method achieves new state of the art results on these datasets, and outperform all recent deep learning approaches to the problem. We introduced the novel concept of Predict and Constrain, which we hope to be further explored in the future and applied to additional application domains. 

The general underlying approach we propose is to consider high order constraints where the value of the constraint is predicted by one network, and another network implements projecting on this constraint. It will be interesting to explore other types of constraints and projections where this is possible from an algorithmic perspective and effective empirically.

\bibliographystyle{icml2018}

\end{document}